\documentclass[conference]{IEEEtran}
\ifCLASSINFOpdf
\else
\fi

\usepackage{multirow}
\usepackage{graphicx}
\usepackage{subfig}
\usepackage{tabularx}
\begin{document}
\setlength{\belowcaptionskip}{0pt}

%
\title{Automatic Detection and Decoding of Photogrammetric Coded Targets}

\author{
\IEEEauthorblockN{Udaya Wijenayake, Sung-In Choi and Soon-Yong Park}
\IEEEauthorblockA{School of Computer Science Engineering\\
Kyungpook National University, South Korea\\
udaya@vision.knu.ac.kr, ellim5th@naver.com, sypark@knu.ac.kr}
}


%


\maketitle

\begin{abstract}
Close-range Photogrammetry is widely used in many industries because of the cost effectiveness and efficiency of the technique. In this research, we introduce an automated coded target detection method which can be used to enhance the efficiency of the Photogrammetry.
\end{abstract}


%
\IEEEpeerreviewmaketitle

\section{Introduction}
Photogrammetry is a technique to determine the geometric properties of an object from a sequence of photographic images, which has a long history since 1950s. Photogram¬metry can be classified into two main categories as Aerial Photogrammetry and Close Range Photogrammetry (CRP) based on the camera location during the image acquisition process. In CRP, the camera is closed to the subject and 3D models, measurements and point clouds are the most common outputs. This technique is used in different application areas, such as modeling and measuring buildings and large engineering structures, accident scenes inspection, mine inspection, quality controlling and film industry.

\begin{figure}[!b]
\begin{center}
\subfloat[]{\label{binary}\includegraphics[width=2.7cm]{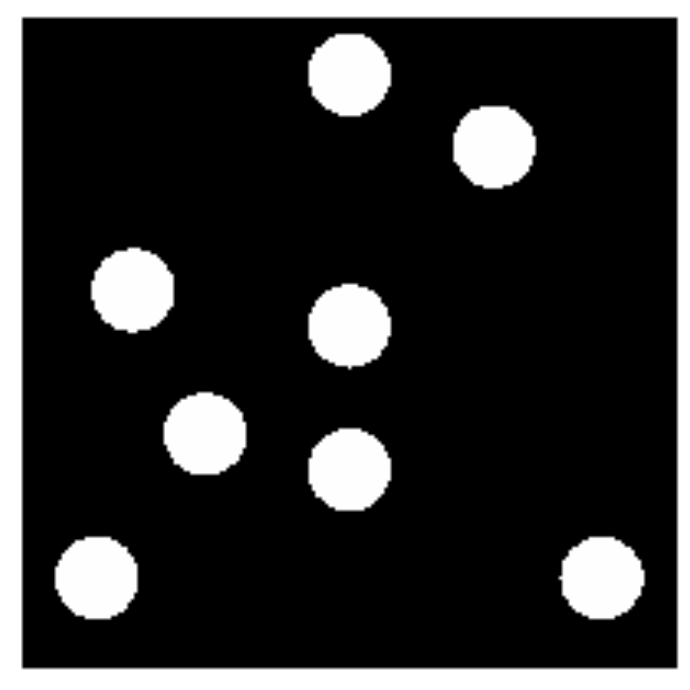}}
\hspace{0.1em}
\subfloat[]{\label{binary}\includegraphics[width=2.7cm]{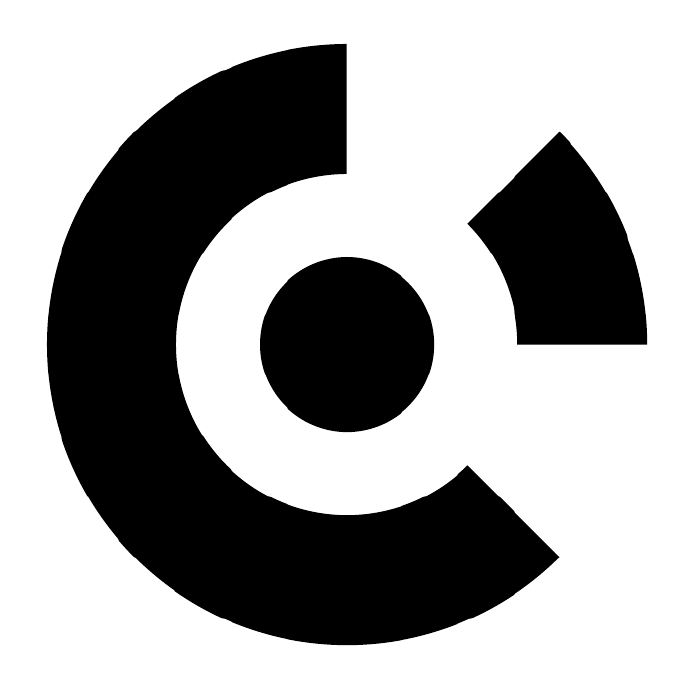}}
\hspace{0.1em}
\end{center}
\caption{Two types of CTs. (a) Dot distributions (b) Concentric rings.}
\label{fig:fig1}
\end{figure}

\section{Coded Targets}
Industrial photogrammetric measurement techniques utilize retro-reflective coded targets (CTs) encoded with a unique identifier to signalize feature points. Two main categories of CTs; concentric rings [1, 2] and dot distributions [3-5] have been introduced so far and two examples are shown in Fig. \ref{fig:fig1}.

Concentric rings are relatively easy to recognize and decode, but gives a limited number of code possibilities. Therefore, it is not suitable for measuring large scale structures which need several hundreds of markers to cover the whole structure. Dot distribution targets solve this problem by allowing a large number of code possibilities, but have difficulties in recognition due to the complexity. 

\begin{figure}[!b]
\centering
\includegraphics[width=0.4\textwidth]{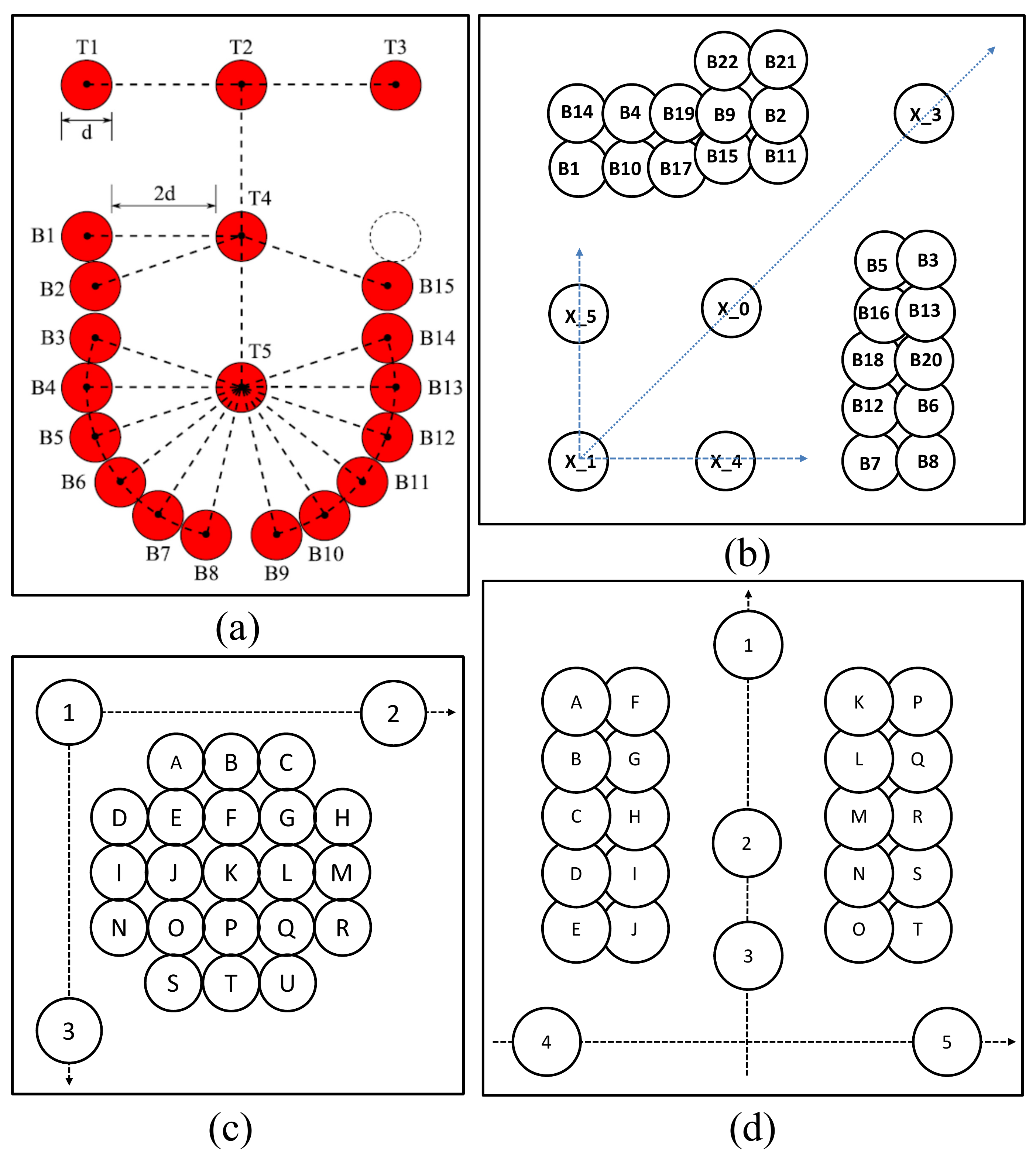}
\caption{Different types of dot distribution CTs [3-5].}
\label{fig:fig2}
\end{figure}

Different dots distribution CTs have been introduced so far and some examples are shown in Fig. \ref{fig:fig2}. CRP systems work with a large number of CTs within an image and 30 to 100 images for one network. Therefore, it is really important to achieve fully automatic target detection to make the system cost and time effective. In this paper, we introduce a fully automatic coded target detection and decoding method based on the CT shown in Fig. \ref{fig:fig2}(b).

\section{Coded Target Detection}
The coded target detection process starts by converting the captured image to gray-scale and removing the image noise using the Gaussian smoothing filter, explained in the Eq. \ref{Eq1}. Then the adaptive inverse thresholding with a Gaussian kernel is employed to convert the smoothed image to a binary image. Adaptive thresholding defines different threshold values for each pixel in the image by examining the surrounding neighborhood pixels as explained in Eq. \ref{Eq2} where $T(x,y)$ is the weighted sum of the neighborhood of $(x,y)$ and $c$ is a user defined constant.

\begin{equation}
g(x,y) = \frac{1}{{2\pi {\sigma ^2}}}{e^{ - \frac{{{x^2} + {y^2}}}{{2{\sigma ^2}}}}}
\label{Eq1}
\end{equation}

\begin{equation}
dst(x,y) = \left\{ {\begin{array}{*{20}{c}}
0\\
{255}
\end{array}} \right.\begin{array}{*{20}{c}}
{if\;src(x,y) > T(x,y) - c}\\
{otherwise}
\end{array}
\label{Eq2}
\end{equation}

Contour detection is applied to the binary image for finding possible circular dots belong to the CTs. After detecting contours, first we remove very large or small contours which are possibly not circular dots by examining the contour size. Then each contour is approximate to a polygon using Douglas-Peucker algorithm and analyze the shape to remove the contours which are not convex and have fewer number of sides. Algorithm introduced in [6] is used to fit each contour to an ellipse and find the center coordinates of circular dots. Finally, these detected circular dots are grouped into possible markers (CTs) using a nearest neighbor search algorithm based on a k-d tree data structure.

\begin{figure}[!b]
\centering
\includegraphics[width=0.4\textwidth]{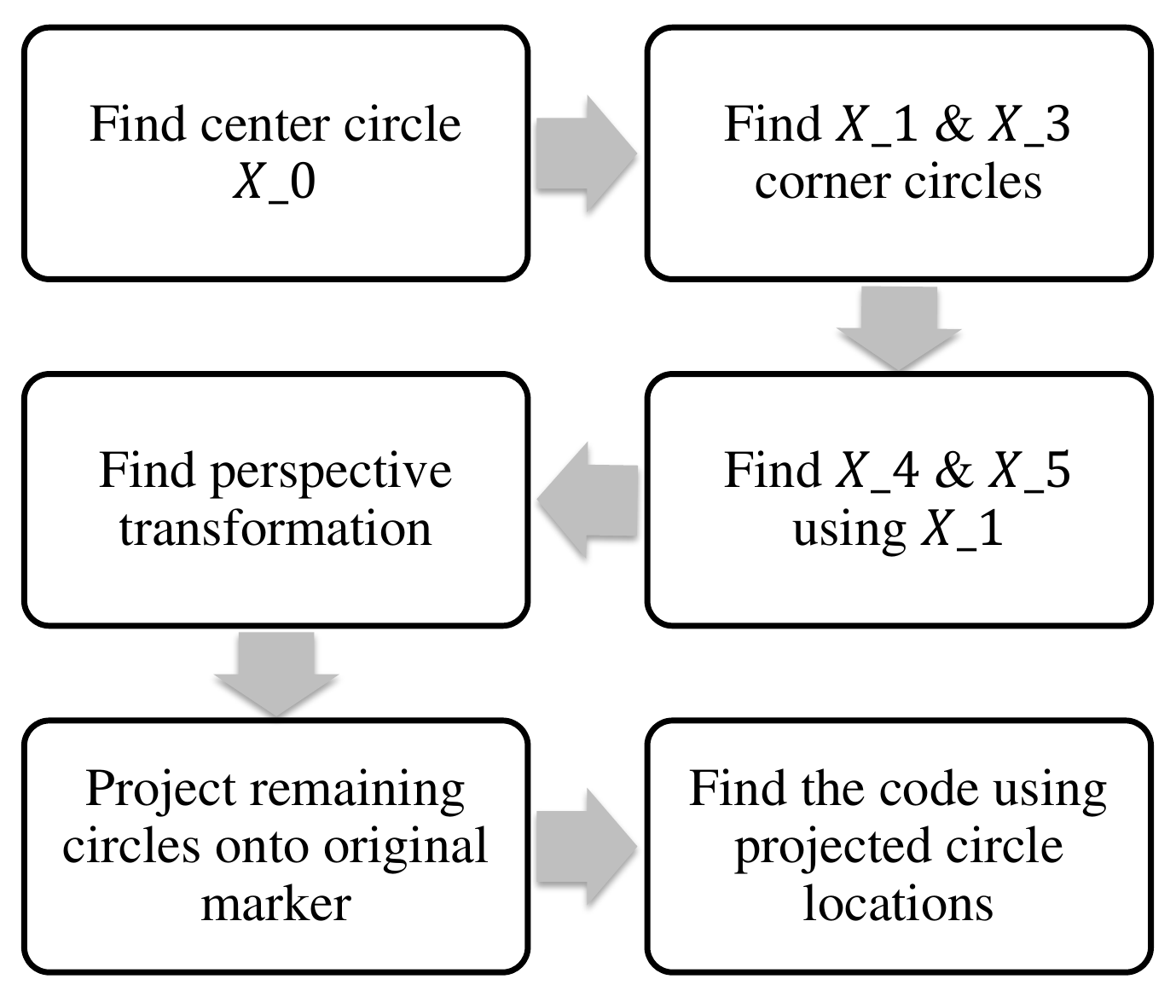}
\caption{Flow chart of the decoding process.}
\label{fig:fig3}
\end{figure} 

\section{Coded Target Decoding}
Each CT use in this research has eight circular dots where five of them are common for all the CTs and define the axis (orientation) of the CT. Remaining three circles can be chosen from 22 possibilities and they define the code-word of the CT. The first step of the decoding process is to identify the five common circles from the detected possible CTs.

We measure the total distance from each circle to every other circles within a CT and select the circle with the lowest total as center circle $x_0$. Then we identify the two corner circles $x_1$ and $x_3$ by finding the two circles which make a straight line with $x_0$. From all the eight circles, only $x_1$, $x_0$ and $x_3$ make a straight line within a CT. $x_1$ and $x_3$ can be differentiated by finding the distance to the $x_0$. To find the $x_4$ and $x_5$, we find the two closest circles to the $x_1$ from the remaining circles. Those two circles should reside on the two sides of the line created by $x_1$ and $x_3$ and can be differentiated as $x_4$ and $x_5$ according to the side they reside. 

After finding the five common circles, the perspective transformation between the original projected CT and the detected one can be calculated. Then the remaining three circles can be projected onto the original marker and the code-word of the CT can be identified using a mask created by original positions of the code circles. Flow chart in Fig. \ref{fig:fig3} summarizes this decoding process.

\begin{figure}[!t]
\centering
\includegraphics[width=0.45\textwidth]{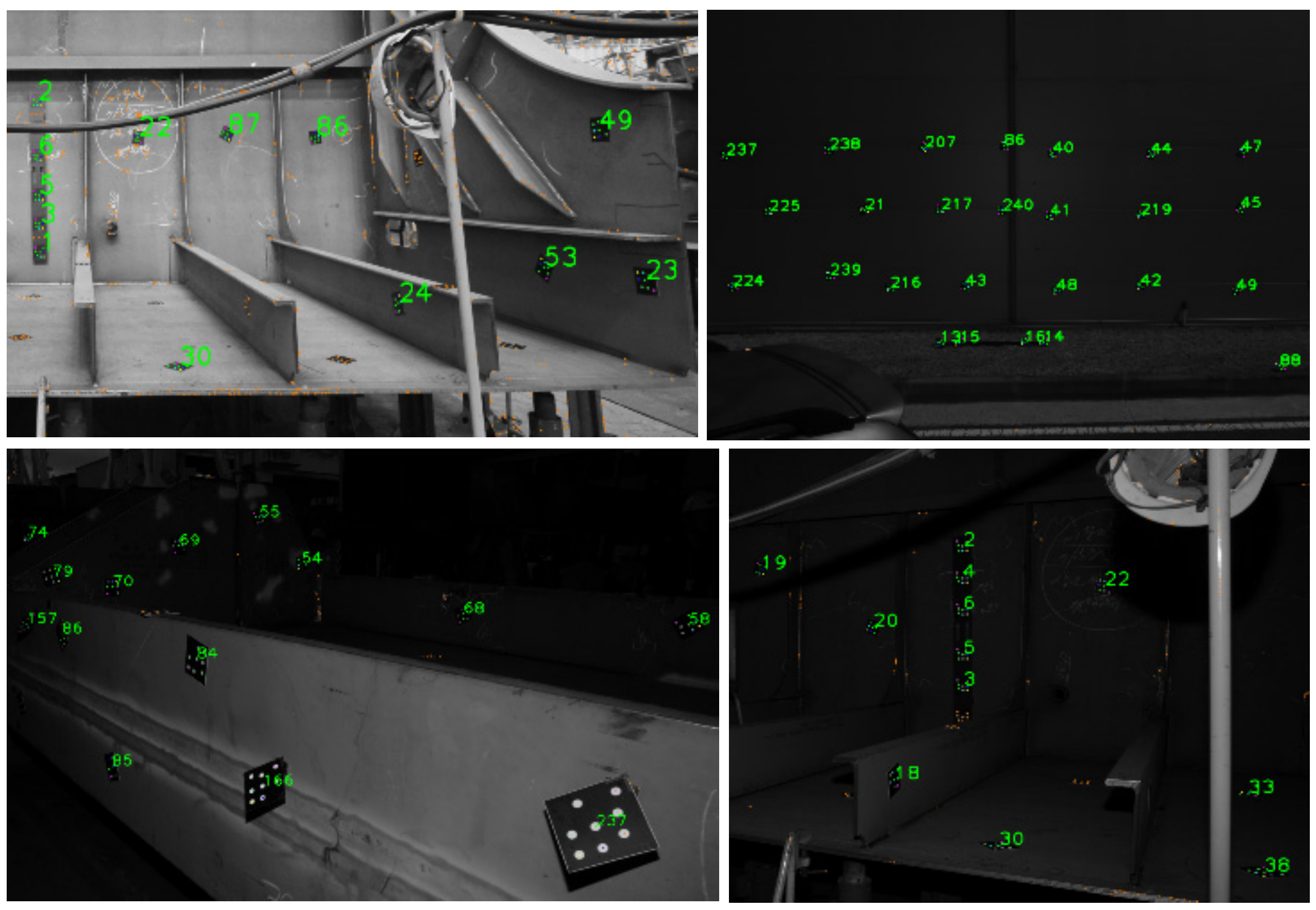}
\caption{Results of the experiments with different structures.}
\label{fig:fig4}
\end{figure}

We test our proposed method with different structures (Fig. \ref{fig:fig4}) and achieve nearly 100\% detection rate when the CTs are nearly orthogonal to the camera view and about 85\% for other cases. For every successful detection of a CT, decoding accuracy is almost 100\% and most of the errors come from the target detection process.

\section{Conclusions}
In an effort to enhance the performance of photogrammetric measurement techniques, in this research, we proposed a fully automated coded target detection and decoding method by using various image processing and vision techniques. As a future work we are planning to integrate this target detection technique with natural feature tracking techniques to develop a more accurate and enhanced photogrammetric system.

\section*{Acknowledgment}

This work was supported by the Industry Core Technology Program granted financial resource from the Ministry of Trade, Industry \& Energy, Republic of Korea (No. 10043897) and Samsung Heavy Industries Co.



%

\end{document}